\documentclass{article}

\usepackage[preprint]{neurips_2022}

\usepackage[utf8]{inputenc} 
\usepackage[T1]{fontenc}    
\usepackage{hyperref}       
\usepackage{url}            
\usepackage{booktabs}       
\usepackage{amsfonts}       
\usepackage{nicefrac}       
\usepackage{microtype}      
\usepackage{xcolor}         
\usepackage{graphicx}
\usepackage{placeins}
\usepackage{rotating}
\usepackage{svg}
\usepackage{amsmath}

\title{Interpretable Polynomial Neural Ordinary Differential Equations}

\author{%
  Colby Fronk 
    \\
  University of California, Santa Barbara\\
  Santa Barbara, CA 93106\\
  \texttt{colbyfronk@ucsb.edu} \\
  \And
  Linda Petzold \\
  University of California, Santa Barbara\\
  Santa Barbara, CA 93106\\
  \texttt{petzold@ucsb.edu} \\
}

\begin{document}

\maketitle

\begin{abstract}
  Neural networks have the ability to serve as universal function approximators, but they are not interpretable and don’t generalize well outside of their training region.  Both of these issues are problematic when trying to apply standard neural ordinary differential equations (ODEs) to dynamical systems.  We introduce the polynomial neural ODE, which is a deep polynomial neural network inside of the neural ODE framework.  We demonstrate the capability of polynomial neural ODEs to predict outside of the training region, as well as to perform direct symbolic regression without using additional tools such as SINDy.  
\end{abstract}

\section{Introduction}

Dynamical systems are mathematical equations written to describe the interactions of quantities that change in time in many science and engineering applications.  Traditionally, mechanistic models describing these systems were obtained from the iterative process of deriving equations from first principles and testing the models with physical experimentation.  The power of mechanistic models comes from the ability to directly explain the system at hand with known physical principles such as the thermodynamics of the system, heat and mass transfer processes, chemical kinetics, and the system's forces.  Reliable predictions can be made about the system behavior in different regimes from mechanistic models due to the scientist's knowledge of the assumptions governing the various first principles and physical laws.  Mechanistic models are preferred by scientists and engineers; however, it can take years to develop a complete and accurate description of a system and we do not know the underlying first principles describing all systems.  \cite{SINDY} sped up the identification of mechanistic models for nonlinear dynamical systems with the method Sparse Identification of Nonlinear Dynamics (SINDy), which is a sparse regression problem of time derivatives obtained by finite difference methods with a library of candidate terms appearing in the dynamical system.  SINDy has had great success with system identification for various applications such as plasma physics \citep{plasma_SINDy}, nonlinear optical communication \citep{nonlinear_optics_SINDy}, biological chemical reaction networks \citep{reaction_networks_SINDy, reactive_SINDy}, and fluid dynamics \citep{PDE_SINDy}.

With the emergence of increased computational power from GPUs and CPUs, along with the exponential growth of the amount of data gathered such as via automated experimentation in chemistry and biology \citep{trends_highthroughput_screening, szymanski2011adaptation}, sensor data from factories \citep{sensor_advances}, satellite and in situ earth observations \citep{satellite}, and large-scale simulations such as computational fluid dynamics (CFD) simulations \citep{calzolari2021deep} and climate models \citep{Randall, esd-12-401-2021}, data-driven models such as deep learning have emerged as a way to process and understand this large amount of data quickly.  

Neural ordinary differential equations (ODEs) are a recent approach to data-driven modeling of time-series data and dynamical systems in which a neural network is used to learn an approximation to an equation governing the dynamics of the system.  Neural ODEs were first introduced in \cite{NeuralODEPaper}'s seminal NeurIPS best paper.  In the last few years, several other types of neural ODEs have emerged such as latent ODEs \citep{latent_ODEs}, Bayesian neural ODEs \citep{bayesianneuralode}, and neural stochastic ODEs \citep{stochastic_neural_ode}.  Models describing the system can be created solely from observed data without the need for expert domain knowledge, which \cite{universal_diffeq} term as the universal differential equation.  These models can then be used to make predictions of what the system will do for unobserved conditions.  Traditional data-driven neural ODE models can be obtained in a few hours as opposed to the many years required to develop complete and accurate mechanistic models for dynamical systems.  Neural ODEs integrate the neural ODE in time to obtain predictions for the observed data, whereas SINDy uses finite difference methods with the data to obtain numerical approximations for the time derivatives, which gives neural ODEs the advantage of having less stringent requirements on the frequency of the observations \citep{SINDy-sampling}.  However, neural ODEs have the well-known major problems that they are not directly interpretable, and they do not make reliable predictions outside of the domain of their training region.  \citep{SINDY-neural} were able to make the neural ODE interpretable by successfully recovering symbolic equations from conventional neural ODEs by using SINDy with predictions for the time derivatives obtained from the trained neural ODE.  

Our work is motivated by the need to make neural ODEs directly interpretable without using additional methods like SINDy after training.  We address this problem by making the case to use directly interpretable neural network architectures inside of the neural ODE framework rather than the conventional neural network based on the standard multilayer perceptron (MLP) \citep{Goodfellow-et-al-2016}. Nonlinear dynamical systems can be complex symbolic expressions.  We chose to address the class of dynamical systems described by polynomials first before tackling more advanced expressions.  Dynamical systems involving polynomials arise in a number of physical systems such as gene regulatory networks \citep{sanguinetti2018gene} and cell signaling networks \citep{gutkind2000signaling} in systems biology, population models in ecology \citep{royle2008hierarchical} and epidemiology \citep{singh2018mathematical}, and atmospheric chemical kinetics \citep{brasseur2017modeling}. In the process of this work, we developed a few deep polynomial neural network architectures, but had the most success with \citep{PiNetPaper}'s $\pi$-net.  We are the first to use any deep polynomial neural network for the purpose of direct symbolic regression, which we demonstrate on a fourth order univariate polynomial.  We are also the first to put a deep polynomial neural network into the neural ODE framework, which we term the polynomial neural ODE, and use the polynomial neural ODE to perform direct symbolic regression on nonstiff dynamical systems described by a polynomials such as the Lotka-Volterra model, Damped Oscillatory System, and Van der Pol model.  In addition, we test the polynomial neural ODE on a model that does not involve polynomials, and demonstrate its effectiveness as another form of a deep polynomial function approximator.  This work, along with future work, will allow the neural ODE to serve as a tool complementary to SINDy for system identification.

\FloatBarrier
\section{Methods}
\FloatBarrier
\subsection{Neural ODEs}

Neural ordinary differential equations (ODEs) learn an approximate ODE, given data for the solution, y(t) \citep{NeuralODEPaper}.  The ODE that we seek to approximate is given by  

\begin{equation}
    \frac{dy(t)}{dt} = f\left(t, y(t), \theta \right),
\end{equation}

\noindent where t is the time, $y(t)$ is the vector of state variables, $\theta$ is the vector of parameters, and $f$ is the ODE model.  For many scientific problems, it can take years to discover the functional form of the ODE described by $f$. Neural ODEs solve this problem by learning an approximation to the dynamics described by $f$ without learning the exact functional equation.  The neural ode, which we denote by $NN$, is a neural network that approximates the model $f$:

\begin{equation}
    \frac{dy(t)}{dt} \approx NN\left(t, y(t), \theta \right).
\end{equation}

\noindent Once the neural ODE has been trained, it is treated exactly the same way as an ODE.  To obtain predictions for $y(t)$, the neural ODE is integrated as an initial value problem (IVP) with an ODE solver.  Traditionally, neural ODEs have used the same architecture as a standard multilayer perceptron (MLP)

\begin{equation}
    NN(x) = (L_1 \circ L_2 \dots \circ L_{l-1} \circ L_l)(x),
\end{equation}

\noindent that is the composition of several neural network layers, $L_i$

\begin{equation}
    L_i(x) = \sigma(x*w_i+b_i),
\end{equation}

\noindent with nonlinear activation function $\sigma$, weights $w_i$, and bias $b_i$ \citep{Goodfellow-et-al-2016}.

\FloatBarrier
\subsection{Polynomial Neural ODEs}

Mathematical models in numerous application areas including gene regulatory networks \citep{sanguinetti2018gene} and cell signaling networks \citep{gutkind2000signaling} in systems biology, population models in ecology \citep{royle2008hierarchical} and epidemiology \citep{singh2018mathematical}, and atmospheric chemical kinetics \citep{brasseur2017modeling} are expressed as differential equations where the right hand side functions $f$ are polynomials.  For this class of problems we present the polynomial neural ODE.  While \cite{Learning-Polynomials} have theoretically and experimentally shown that conventional feedforward MLPs work as universal approximators for polynomials, we will make the case for using polynomial neural ODEs instead of conventional neural ODEs for this application space. 

\begin{figure}[h!]
    \centering
    \includegraphics[width=\textwidth]{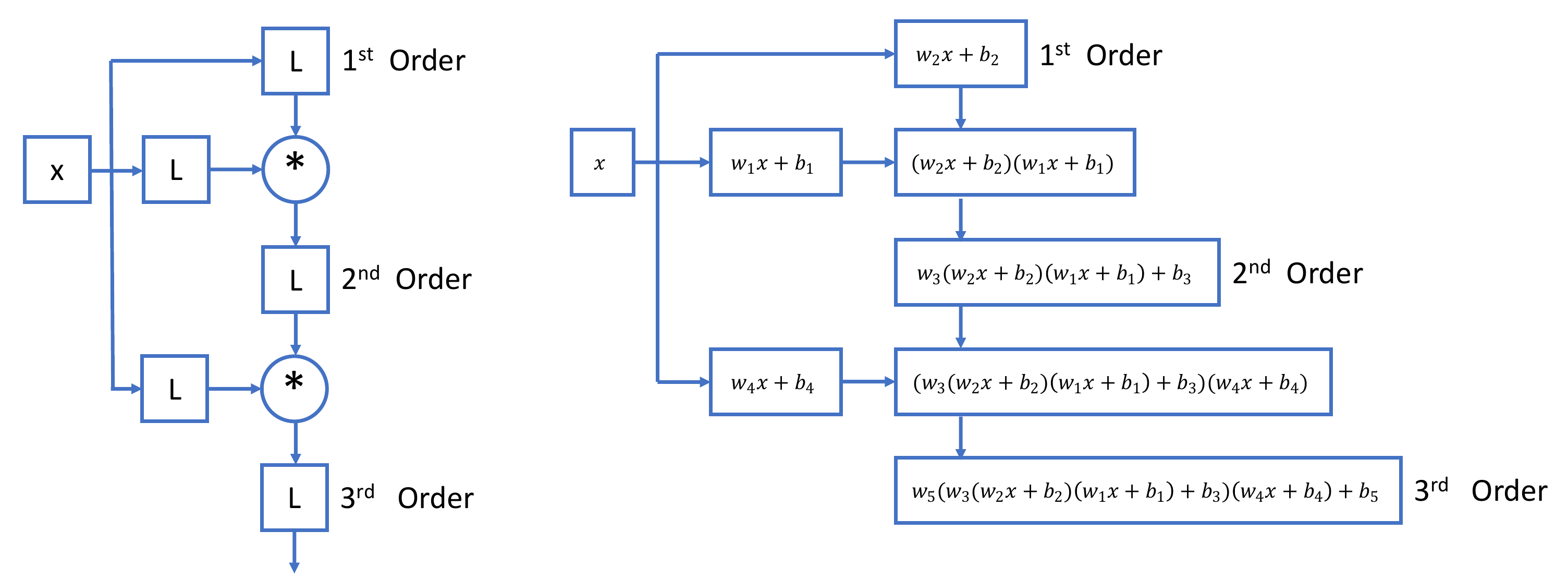}
    \caption{The network architecture of \cite{PiNetPaper}'s $\pi$-net V1 is shown (left) alongside a worked example of a one layer input with variable name $x$ (right).  The boxes labeled L represent standard linear layers, while the circles with the $*$ symbol represent a layer that is the Hadamard product of the layer's inputs.  There are no standard activation functions such as tanh or ReLU in this network architecture, which makes the architecture directly interpretable.}
    \label{fig:PiNetV1_arch}
\end{figure}

Polynomial neural networks are function approximators in which the output layer is expressed as a polynomial of the input layer.  
There are several types of polynomial neural networks.  Designing polynomial neural networks that can be trained easily without an explosion of parameters is still an active area of research \citep{PiNetPaper, FAN2020383, du2018power, liang2016deep}. In this paper, we use the $\pi$-net architecture from \citet{PiNetPaper}.    $\pi$-nets were specifically designed to use skip connections to avoid the combinatorial explosion in the number of parameters of polynomial activation functions, which make the network harder to train.  Essentially, these networks learn tensor decompositions of the polynomials.  For our work we use $\pi$-net V1, for which the architecture is shown in Fig. \ref{fig:PiNetV1_arch}.  The architecture is centered around Hadamard Products \citep{horn1994topics} of linear layers without activation functions to form higher order polynomials.  

\begin{figure}[!ht]
    \centering
    \includegraphics[width=\textwidth]{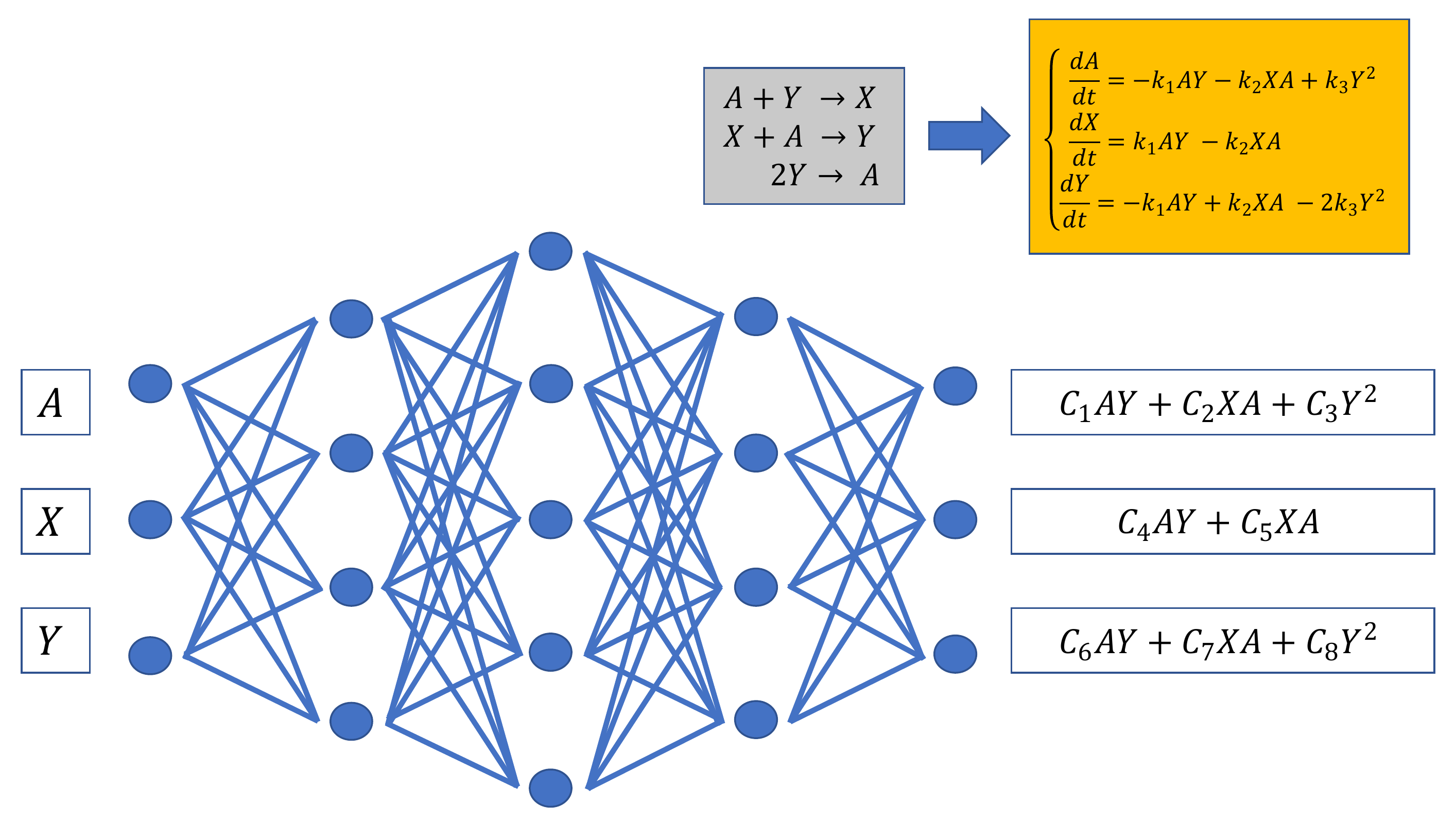}
    \caption{Example polynomial neural ODE for a chemical reaction system with molecules A, X, and Y.  The neural network outputs a polynomial transformation of the input, which are the concentrations of the chemical species.}
    \label{fig:NN_visual}
\end{figure}

Conventional neural networks are not interpretable, due to the complex arrangement of nonlinear activation functions.  \cite{SINDY-neural} have had success using \cite{SINDY}'s Sparse Identiﬁcation of Nonlinear Dynamics (SINDy) to recover symbolic forms of neural ODEs following training. However, the polynomial neural ODE architecture has a unique advantage.  Since the output layer is a direct mapping of the input in terms of tensor and Hadamard products, symbolic tensor math can be used to obtain a direct polynomial form of the polynomial neural network without using additional tools such as SINDy.  We use SymPy, the Python library for symbolic mathematics, to obtain a symbolic form of the polynomial neural ODE following training \citep{SymPy}.  Since we expect to see a plethora of new neural network architectures for symbolic regression beyond polynomials, we anticipate the need for compatibility of symbolic computation with software such as PyTorch, TensorFlow, and JAX, which would avoid the need to extract the network architecture, weights, and biases into SymPy following training to obtain a symbolic equation.

\FloatBarrier
\subsection{Training Neural ODEs}

Prior to training the neural ODE, the architecture is defined and the parameters in the network are initialized.  For all of our neural ODEs, we initialized our weights and biases with a normal distribution with mean of 0. We used a standard deviation of 0.00005 and 0.01 for the conventional and $\pi$-net V1 neural ODEs respectively.  

When training a neural ODE, the goal is to fit the neural ODE to observed data for the state variables, $y_{obs}$, as a function of time.  The neural ODE is integrated with an ODE solver to obtain predictions for $y_{obs}$, which we will call $y_{pred}$.  We used gradient descent to minimize the normalized MSE loss $L(y_{obs}, y_{pred})$ between $y_{obs}$ and $y_{pred}$: 

\begin{equation}
    L(y_{obs}, y_{pred}) = \frac{1}{N} \sum_i^N \left (\frac{y_{pred, i} - y_{obs, i}}{y_{scale}}\right)^2,
\end{equation}

\noindent where $y_{scale}=\mid y_{max} - y_{min} \mid$ is the scale factor used to normalize the values in the loss function.  We batched our data into $N-1$ samples consisting of IVPs between 2 adjacent known data points.  For each iteration of gradient descent, we simultaneously solve the $N-1$ initial value problems forward in time to the next observed data point explicitly using the fourth order explicit Runge–Kutta–Fehlberg method \citep{fehlberg1968classical}. Since we are only working with nonstiff ODEs, we are able to use an explicit discretization method to solve the neural ODEs.  The advantage of using this method to solve the neural ODE is efficient direct backpropagation through the explicit ODE scheme, which the popular continuous time sensitivity adjoint method from \cite{NeuralODEPaper} lacks.

\FloatBarrier
\section{Results}
\FloatBarrier
\subsection{Univariate 4th Order Polynomial}

Prior to looking at any dynamical systems with neural ODEs, the ability of $\pi$-net V1 to learn basis polynomials was first tested.  For the test case, we used the following fourth order univariate function:

\begin{equation}
    f(x) = 3 x^4 + 16 x^3 + 5 x^2
\end{equation}

\noindent The training data for the x-values consisted of 20 uniformly spaced data points in the range -5.3 to 2.2.  The values of $f(x)$ corresponding to the values of $x$ were obtained by directly substituting the $x$-values into the function.

Two neural networks were trained using the training data: (1) a conventional neural network with 5 layers consisting of 1x100x100x100x1 neurons in each layer with tanh activation functions and (2) a $\pi$-net V1 polynomial neural network that outputs fourth order polynomials.  Each neural network was trained a total of 5 times and the best network was chosen based on the normalized MSE test loss, $L(y_{obs}, y_{pred})$,  as well as the visual fit of the neural network's predictions against the known data.  

\begin{figure}
    \centering
    \includegraphics[width=\textwidth]{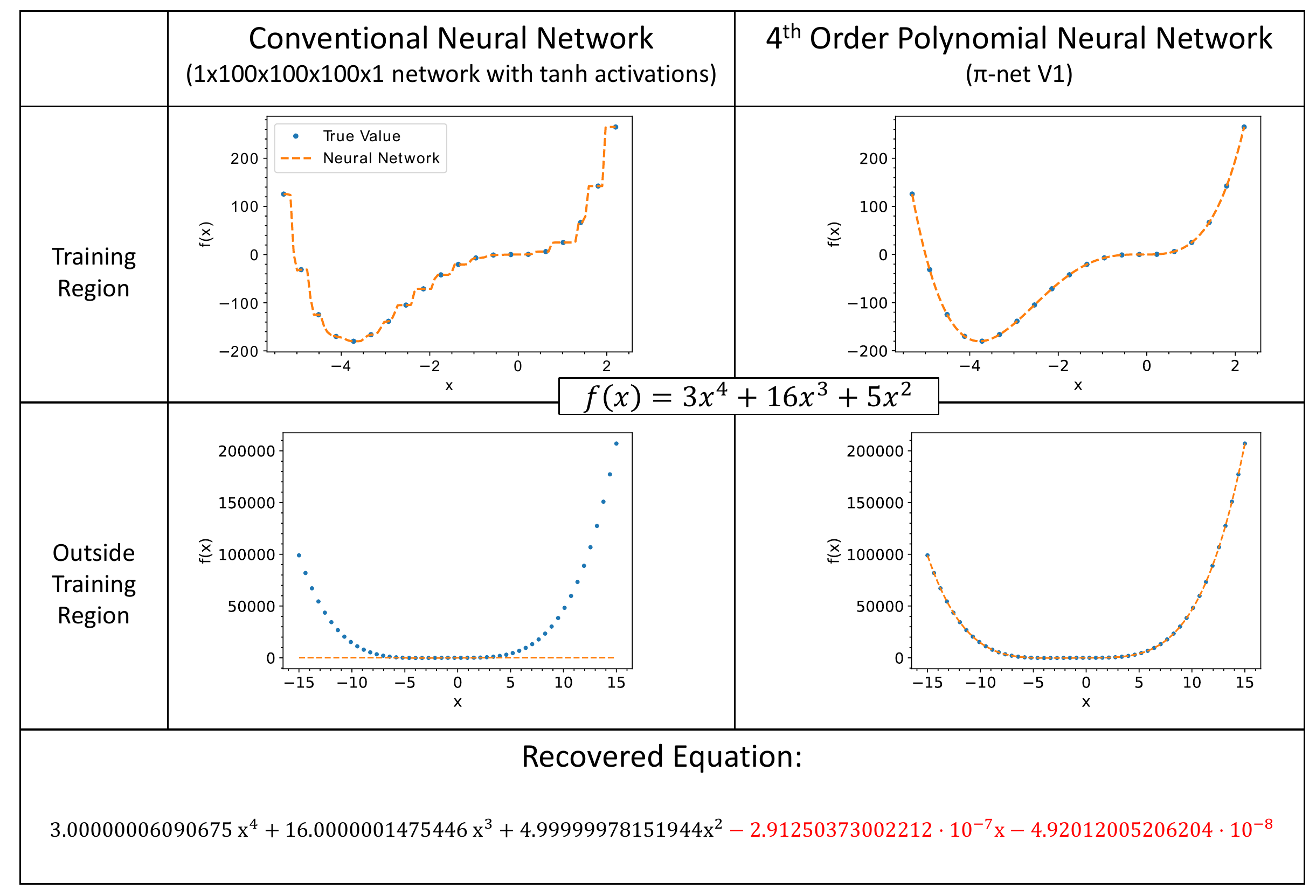}
    \caption{Predictions inside and outside of the training region for a conventional neural network with tanh activation functions, and a fourth order polynomial neural network after learning the fourth order polynomial.  The recovered symbolic equation from the polynomial neural network is shown below the figures.  The red terms indicate additional terms.}
    \label{fig:1D-example}
\end{figure}

Results from the  neural networks with the best fit are shown in Fig \ref{fig:1D-example}.  While not shown, conventional neural networks with ReLU activation functions produced similar results.  The fourth order $\pi$-net V1 polynomial neural network reliably reproduced the same model, due to the polynomial constraint defined by its architecture, whereas there was some variability in the final conventional neural network model.  The conventional neural network exhibits Gibbs phenomenon, an oscillatory behavior around the observed data usually caused by discontinuities, which is typical for approximation functions such as Fourier series, orthogonal polynomials, splines, and wavelets \citep{GibbsPhenom, jerri2013gibbs}.  Since a neural network is another type of approximation function, this behavior is not surprising.  Gibbs phenomenon was found every time the conventional neural network was fit to the data. The model in Fig \ref{fig:1D-example} was chosen because it exhibited the least amount of Gibbs phenomenon.  Additionally, we observed that ReLU activation functions produce Gibbs phenomenon to a lesser extent than tanh activation functions.  We suspect that this is because tanh functions bound the output between -1 and 1, whereas ReLU functions bound the output between 0 and $\infty$, which leads to fewer discontinuities. 

Neural networks are known to be poor at making predictions outside of their training region, which Fig \ref{fig:1D-example} also shows.  When the prediction range is slightly extended, the conventional neural network is completely unable to make an accurate prediction, whereas the polynomial neural network makes accurate predictions in this range because it learned the functional form of the data.  

The main advantage of polynomial neural networks is the ability to directly obtain a symbolic representation of the network.  As shown in Fig. \ref{fig:1D-example}, coefficients of the polynomial accurate to 7 significant figures were directly predicted from 20 observations.  After performing equation recovery, the extra terms can be systematically removed to simplify the model.  In this case, the red terms are small enough to be dropped after further analysis by the modeler.

\FloatBarrier
\subsection{Lotka-Volterra Deterministic Oscillator}

Our first demonstration of a polynomial neural ODE is on the deterministic Lotka-Volterra ODE model, which describes the predator-prey population dynamics of a biological system \citep{Lotka1925, volterra1926variazioni}.  When written as a set of first order nonlinear ODEs, the model is given by

\begin{align} \label{lotka-volterra-equations}
    \frac{dx}{dt} = 1.5 x - x y, \\
    \frac{dy}{dt} = -3 y + x y,
\end{align}

\noindent with initial conditions $x=1$ and $y=1$.
Since the problem is nonstiff, we generate our training data by integrating the IVP with SciPy and torchdiffeq using DOPRI5, a fourth order embedded method in the Runge–Kutta family of ODE solvers, with the default settings at 200 points uniformly spaced in time between 0 and 10 \citep{2020SciPy-NMeth, Chen_torchdiffeq_2021, dopriref}. As discussed in the methods section, we batch our data into 199 training samples consisting of IVPs between 2 adjacent known data points, and simultaneously solve the 199 IVPs during each epoch using our own code for the fourth order explicit Runge–Kutta–Fehlberg method, which allows us to directly perform backpropagation through the ODE discretization scheme \citep{fehlberg1968classical}.  

Four neural networks were trained using the training data: (1) a conventional neural network with 5 layers consisting of 2x50x50x50x2 neurons in each layer with tanh activation functions and (2) three separate $\pi$-net V1 polynomial neural networks of degree two, three, and four. 

\begin{figure}
    \centering
    \includegraphics[width=\textwidth]{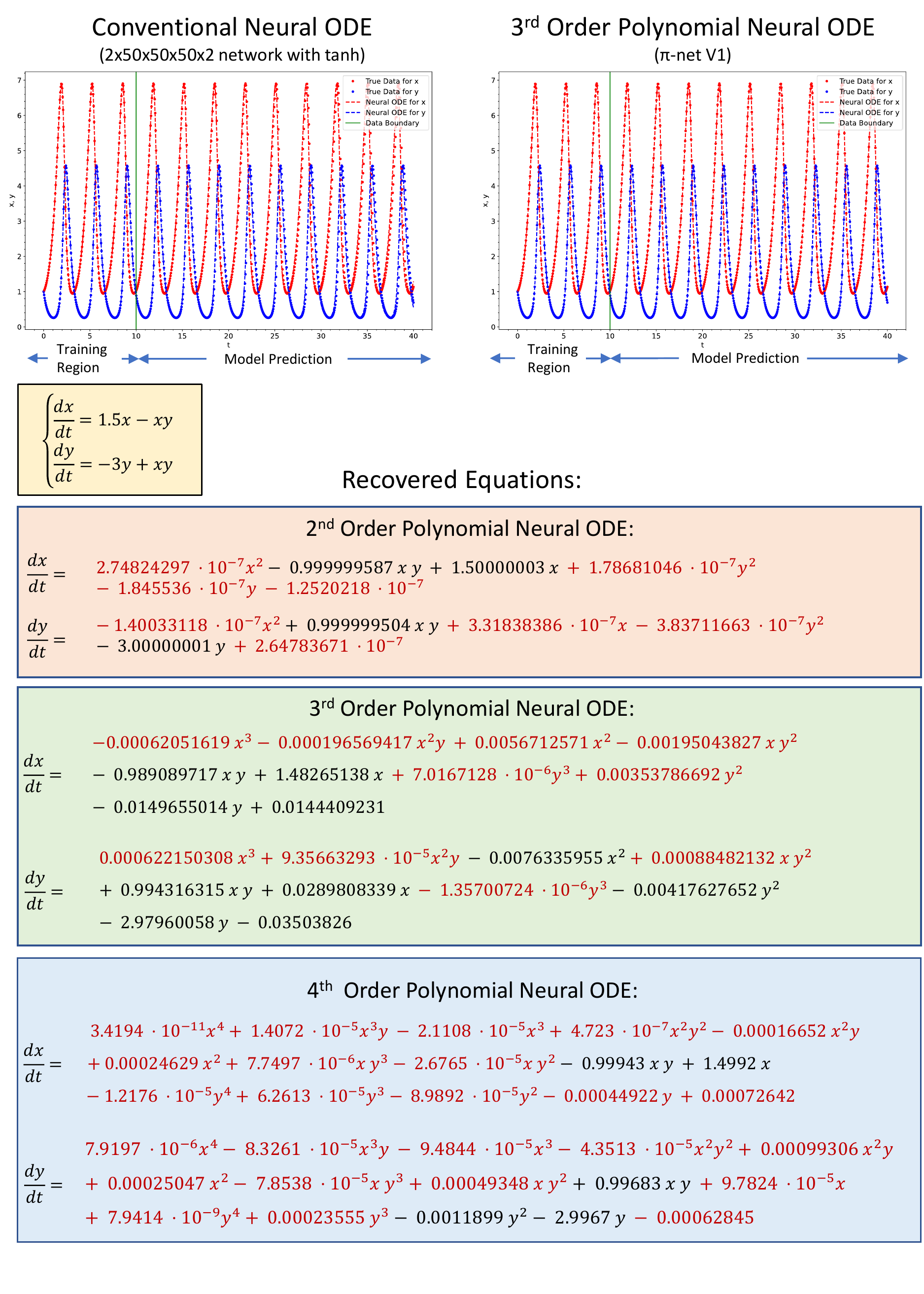}
    \caption{The conventional neural ODE and the $\pi$-net V1 polynomial neural ODE were integrated from the initial conditions to a time of 40, to test the performance of the neural ODEs.  The vertical green line indicates where the training data ends and the neural ODE begins making a prediction.  The symbolic form of the $\pi$-net V1 polynomial neural network was obtained for each of the degrees shown.  The correct equation is shown above the recovered equations.  The red terms indicate terms that can be dropped.}
    \label{fig:PiNet-Lotka-Voltera-Full}
\end{figure}

\begin{figure}
    \centering
    \includegraphics[width=\textwidth]{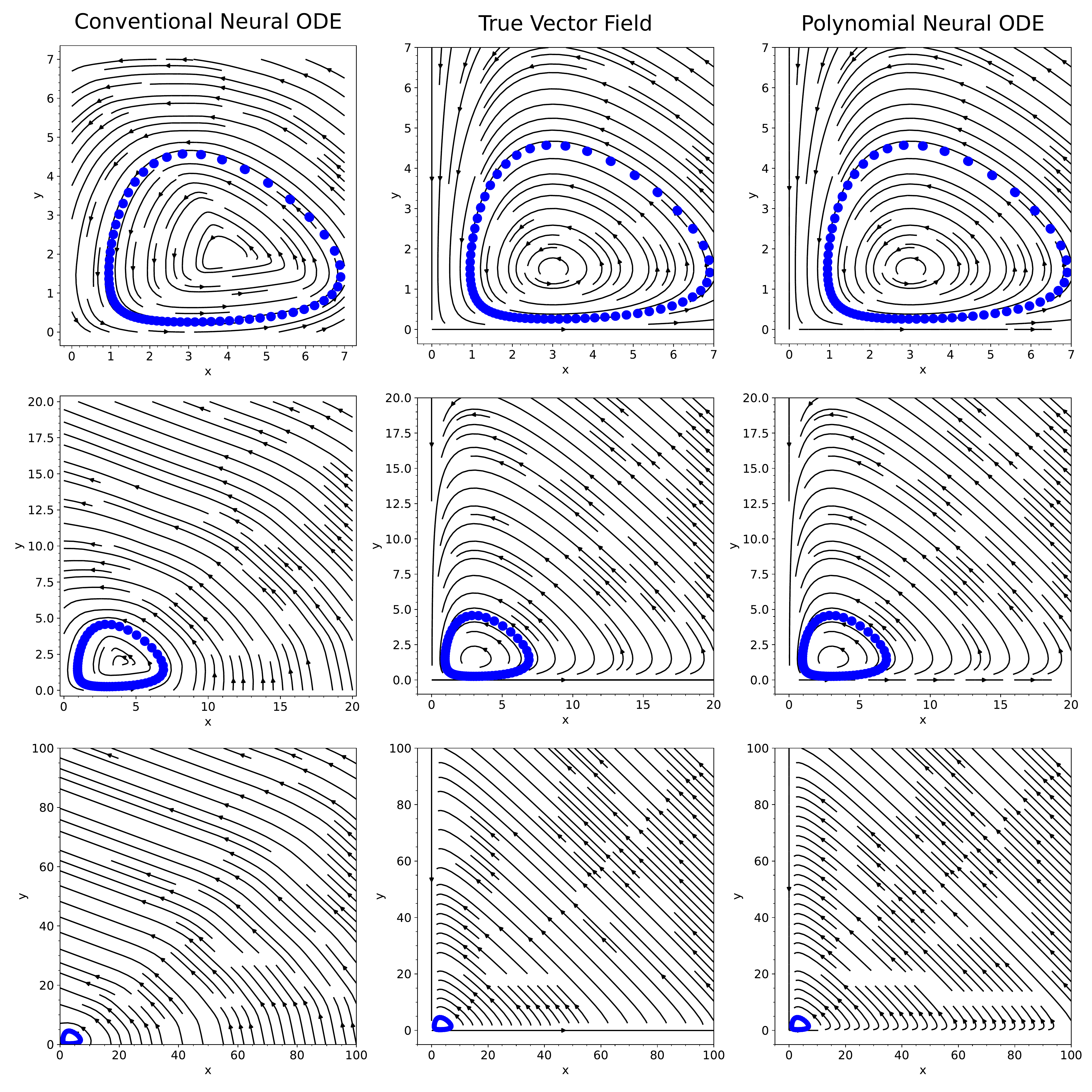}
    \caption{The vector fields of the time derivatives of variables x and y are plotted for the conventional neural ODE with tanh activations (left), the true ODE equation (middle), and the second order $\pi$-net V1 polynomial neural ode (right). The top row shows the training region, while the remaining rows show the region outside of the training region.  The blue circles mark the training data.}
    \label{fig:PiNet-Lotka-Volterra-VectorField}
\end{figure}

Following training, the performance of the neural networks was tested by integrating the neural ODE from the initial conditions up to a time value of 40.  As shown in Fig \ref{fig:PiNet-Lotka-Voltera-Full}, both the conventional neural ODE and the fourth order polynomial neural ODE are able to accurately predict the trajectory of the dynamical system beyond the training region for the same initial values as the training data.  

Fig \ref{fig:PiNet-Lotka-Voltera-Full} doesn't offer much insight about how the conventional neural ODE differs from the polynomial neural ODE, so the vector fields of the time derivatives of x and y were also plotted.  As shown in Fig \ref{fig:PiNet-Lotka-Volterra-VectorField}, the second order $\pi$-net V1 neural ODE accurately learns the true vector field for the training region, as well as outside of the training region.  The conventional neural ODE learns a close approximation to the training region's vector field, which is why it was able to accurately predict the trajectory of the dynamical system beyond the training region for the same initial values as the training data, as shown in Fig \ref{fig:PiNet-Lotka-Voltera-Full}.  However, it doesn't learn an accurate enough approximation to make predictions for observations outside of its training region.  This is clear just by looking at the difference in vector fields for the region of space outside of the training region.  On the other hand, the polynomial neural ODE has no problem with the same task. 

To show that the $\pi$-net V1 polynomial neural ODE learns the same equation regardless of the degree of the polynomial output, three separate polynomial neural ODEs were trained with degrees of two, three, and four.  Following training, the symbolic form of each of the polynomial neural ODEs was obtained, as shown in Fig \ref{fig:PiNet-Lotka-Voltera-Full}.  The second, third, and fourth order polynomial neural ODEs were able to recover 6, 2, and 3 significant digits for the coefficients respectively.  The values of the coefficients belonging to the terms not found in the original ODE differ each time, which serves as a clue to the modeler that these terms can be dropped from the final equation.

\FloatBarrier
\subsection{Damped Oscillatory System}

Our next demonstration of a polynomial neural ODE is on the deterministic damped oscillatory system, a popular toy problem for Neural ODEs \citep{NeuralODEPaper, CollocationNeuralODE}.  Damped oscillations are common in engineering, physics, and biology \citep{doi:10.1080/00107514.2011.644441, karnopp1990system}.  For our work, we refer to the following model as the damped oscillator:

\begin{eqnarray} \label{eq:damped-oscillator}
    \frac{dx}{dt} &=& -0.1 x^3 - 2 y^3  \nonumber \\
    \frac{dy}{dt} &=&  2 x^3 - 0.1 y^3, 
\end{eqnarray}

\noindent with initial conditions $x=1$ and $y=1$.  Since the problem is nonstiff, we generate our training data by integrating the IVP with SciPy and torchdiffeq using DOPRI5, a fourth order embedded method in the Runge–Kutta family of ODE solvers, with the default settings at 100 points uniformly spaced in time between 0 and 25 \citep{2020SciPy-NMeth, Chen_torchdiffeq_2021, dopriref}. As discussed in the methods section, we batch our data into 99 training samples consisting of IVPs between 2 adjacent known data points, and simultaneously solve the 99 IVPs during each epoch using our own code for the fourth order explicit Runge–Kutta–Fehlberg method, which allows us to directly perform backpropagation through the ODE discretization scheme \citep{fehlberg1968classical}.  

Three neural networks were trained using the training data: (1) a conventional neural network with 5 layers consisting of 2x50x50x50x2 neurons in each layer, with tanh activation functions, and (2) two separate $\pi$-net V1 polynomial neural networks of degree three and four.  Previous work has demonstrated that this damped oscillatory system is challenging for traditional neural ODEs to learn \citep{NeuralODEPaper, CollocationNeuralODE}.  Several authors have made the model easier to learn with neural ODEs by making the first layer of the neural network a cubic function; however, this approach requires a priori knowledge that the functional form of the equation is cubic \citep{Chen_torchdiffeq_2021, CollocationNeuralODE}.  Since the traditional neural network is supposed to learn the dynamics on its own, we do not use this approach for any of our neural ODEs.  

\begin{figure}
    \centering
    \includegraphics[width=\textwidth]{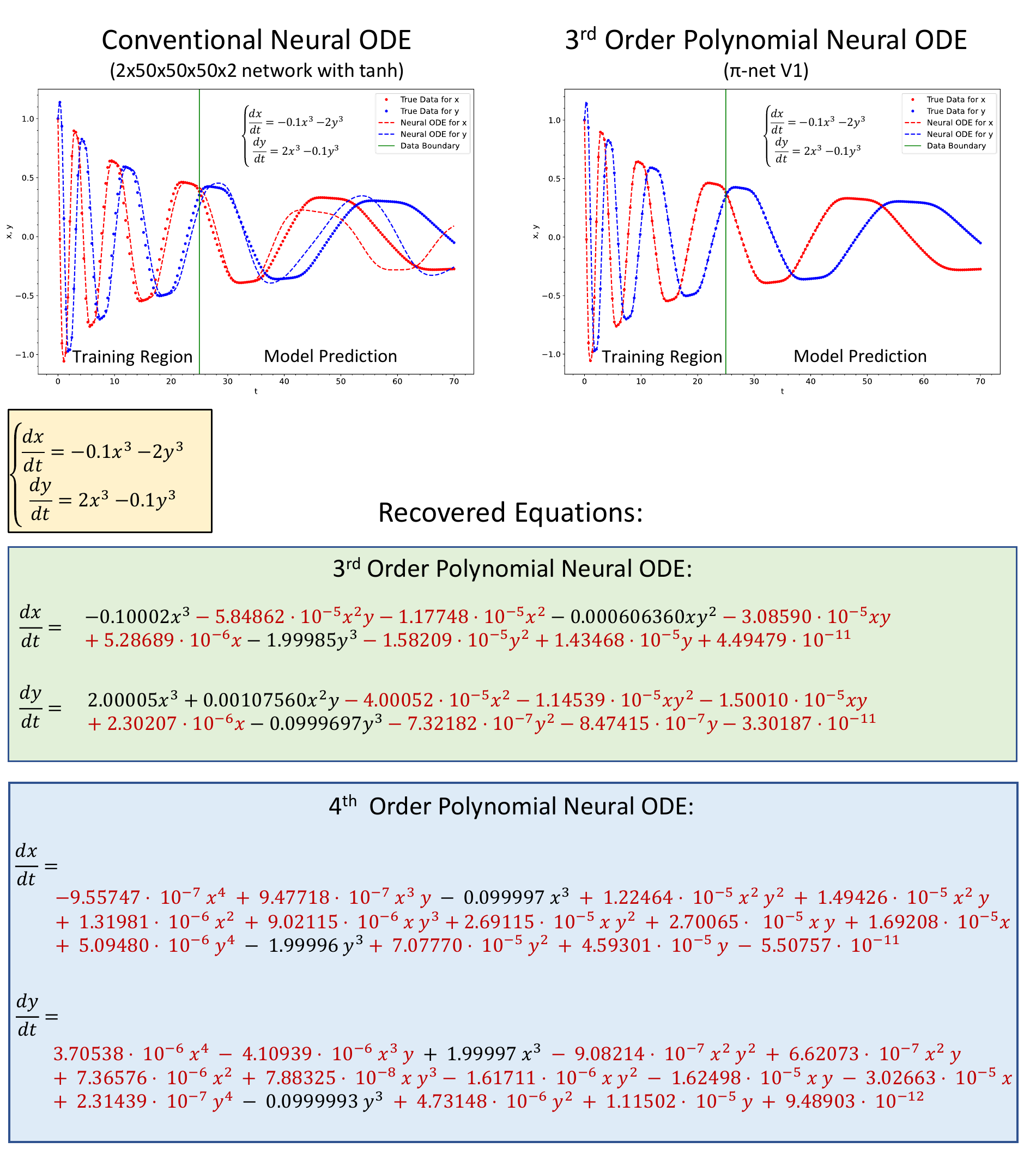}
    \caption{The conventional neural ODE and the $\pi$-net V1 polynomial neural ODE were integrated from the initial conditions to a time of 70, to test the performance of the neural ODEs.  The vertical green line indicates where the training data ends and the neural ODE begins making a prediction.  The symbolic forms of the polynomial neural ODEs were also obtained for each of the degrees shown.  The red terms indicate additional terms that can be dropped.}
    \label{fig:PiNet-Damped-Oscillator-Full}
\end{figure}

\begin{figure}
    \centering
    \includegraphics[width=\textwidth]{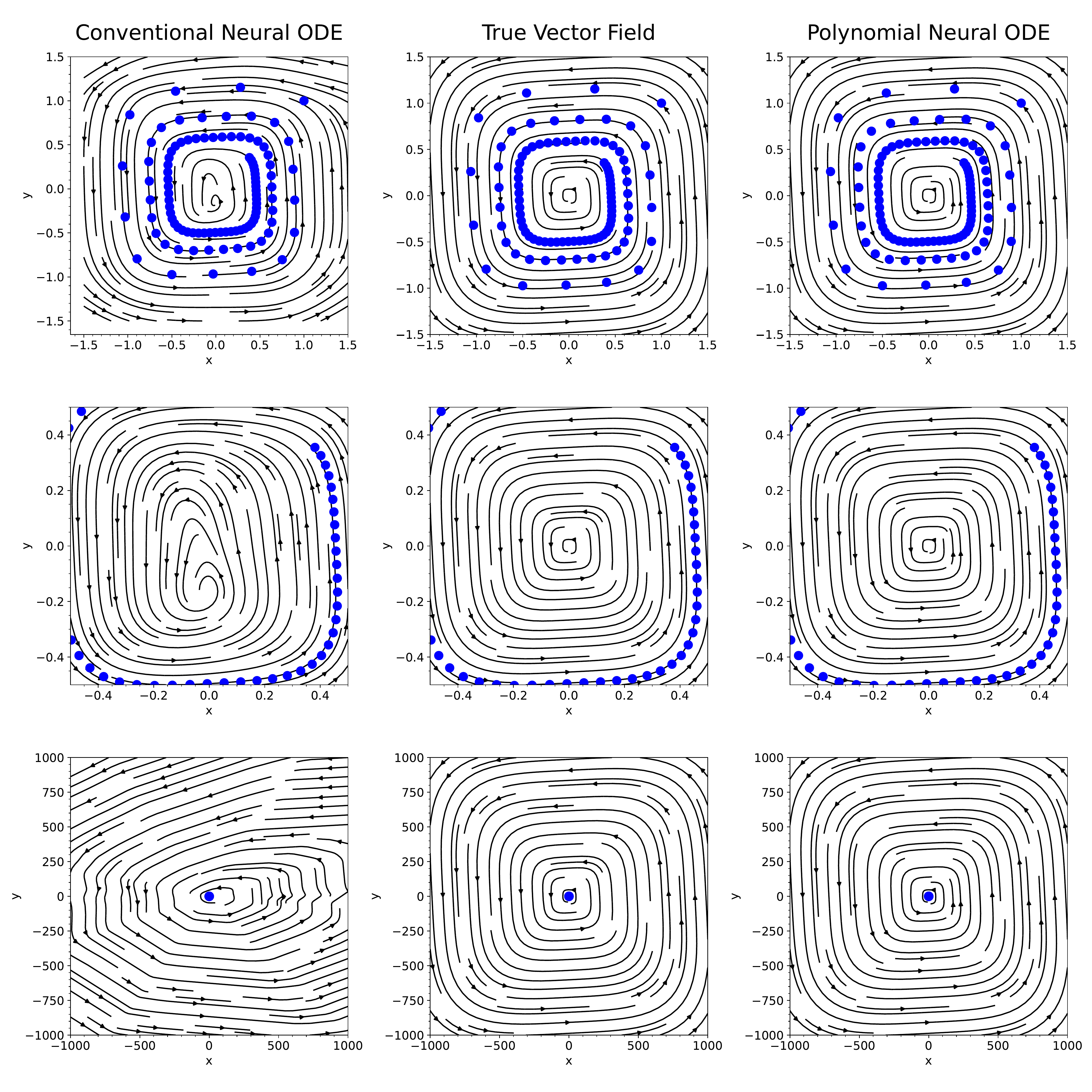}
    \caption{The vector fields of the time derivatives of variables x and y are plotted for the conventional neural ODE with tanh activations (left), the true ODE equation (middle), and the third order $\pi$-net V1 polynomial neural ode (right). The top row shows the entire training region, the middle row shows a smaller subset of the training region where oscillations dampen, and the bottom row shows the region outside of the training region.  The blue circles mark the training data.}
    \label{fig:PiNet-Damped-Oscillator-VectorField}
\end{figure}

Following training, the performance of the neural networks was tested by integrating the neural ODE from the initial conditions up to a time value of 70.  As shown in Fig \ref{fig:PiNet-Damped-Oscillator-Full}, the third order polynomial neural ODE is able to accurately predict the trajectory of the dynamical system beyond the training region, for the same initial values as the training data.  The identical match of vector fields between the third order polynomial and that of the true ODE model, as shown in Fig \ref{fig:PiNet-Damped-Oscillator-VectorField}, indicates why the polynomial neural network is able to accurately predict the trajectory.  

In contrast, the conventional neural ODE is able to make an accurate prediction only up to about 15 time units past the training region.  The conventional neural network's predictions do not preserve the general shape of the solution to the ODE, whereas the polynomial constraint on the polynomial neural ODE ensures that the functional shape is preserved for the prediction task.  Looking at the vector field for the conventional neural ode between -0.5 and 0.5, which is the zoomed in portion of the training region,  since the training data does not have many values in this range, the neural ode was unable to learn the center of the vector field's spiral well.  We observe and hypothesize that this is the reason why the conventional neural ODE is known to struggle with learning the damped oscillator model: the errors from the center of the spiral increase over the trajectory as it dampens.  Additionally, the zoomed out vector field in the range -1000 to 1000 shows that the conventional neural ODE's vector field undergoes major distortion outside of its training region.  This effect is not observed in the polynomial neural network.  

Fig \ref{fig:PiNet-Damped-Oscillator-Full} shows the symbolic equations recovered from the third and fourth order polynomial neural ODEs. Five significant figures were recovered from both of the neural odes, and the remaining coefficients are small enough to be dropped.  

\FloatBarrier
\subsection{Van der Pol Deterministic Oscillator}

Our final demonstration of the performance of polynomial neural ODEs is on the Van der Pol oscillator, a nonconservative oscillator with nonlinear damping \citep{VanderPolModel}.  The Van der Pol equation describes several processes of relaxation-oscillations in the physical and biological sciences.  For example, it has been used to model action potentials of neurons, tectonic plates in a geological fault, and oscillations of the left and right vocal cords during speech \citep{FITZHUGH1961445, 4066548, doi:10.1142/S0218127499001620, doi:10.1121/1.4798467}. The Van der Pol oscillator is described by the following second order ordinary differential equation:

\begin{equation}
    \frac{d^2x}{dt^2}-\mu(1-x^2) \frac{dx}{dt}+x= 0
\end{equation}

\noindent and can be rewritten as a system of first order ODEs:

\begin{eqnarray}
    \frac{dx}{dt} &=& y \nonumber \\
    \frac{dy}{dt} &=& \mu y - \mu x^2 y - x.
\end{eqnarray}

We chose to assign $\mu=5$ and use initial conditions $x=2$ and $\frac{dx}{dt}=y=0$.  Since the problem is nonstiff, we generate our training data by integrating the IVP with SciPy and torchdiffeq using DOPRI5, a fourth order embedded method in the Runge–Kutta family of ODE solvers, with the default settings at 200 points uniformly spaced in time between 0 and 25 \citep{2020SciPy-NMeth, Chen_torchdiffeq_2021, dopriref}. As discussed in the methods section, we batch our data into 199 training samples consisting of IVPs between 2 adjacent known data points, and simultaneously solve the 199 IVPs during each epoch using our own code for the fourth order explicit Runge–Kutta–Fehlberg method, which allows us to directly perform backpropagation through the ODE discretization scheme \citep{fehlberg1968classical}.  Three neural networks were trained using the training data: (1) a conventional neural network with 5 layers consisting of 2x50x50x50x2 neurons in each layer with tanh activation functions, and (2) two separate $\pi$-net V1 polynomial neural networks of degree three and four.

\begin{figure}
    \centering
    \includegraphics[width=\textwidth]{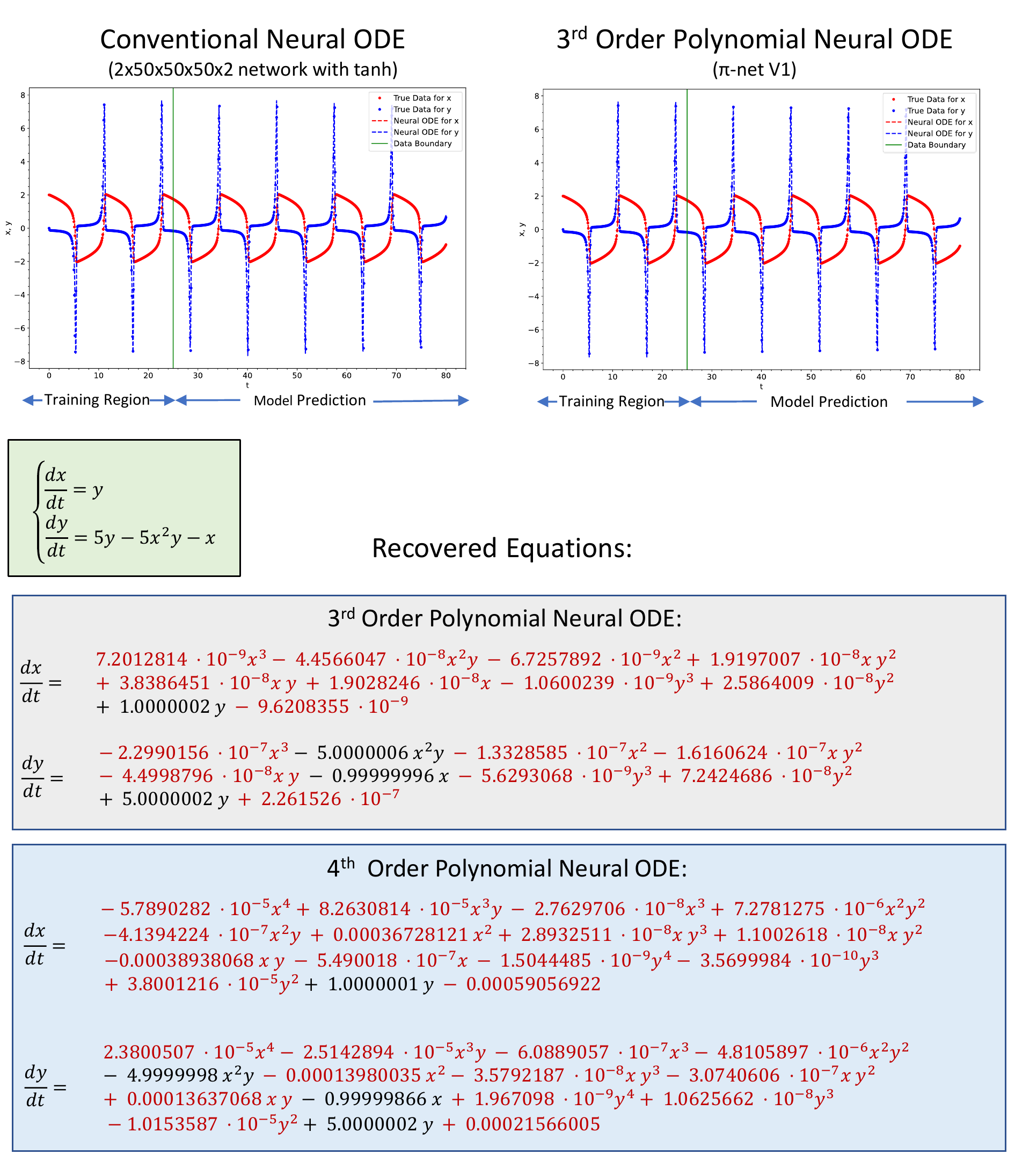}
    \caption{The conventional neural ODE and the $\pi$-net V1 polynomial neural ODE were integrated from the initial conditions to a time of 80 to test the performance of the neural ODEs.  The vertical green line indicates where the training data ends and the neural ODE begins making a prediction. The symbolic forms of the polynomial neural ODEs were also obtained for each of the degrees shown.  The red terms indicate additional terms that can be dropped.}
    \label{fig:PiNet-Van-der-Pol-Full}
\end{figure}

\begin{figure}
    \centering
    \includegraphics[width=\textwidth]{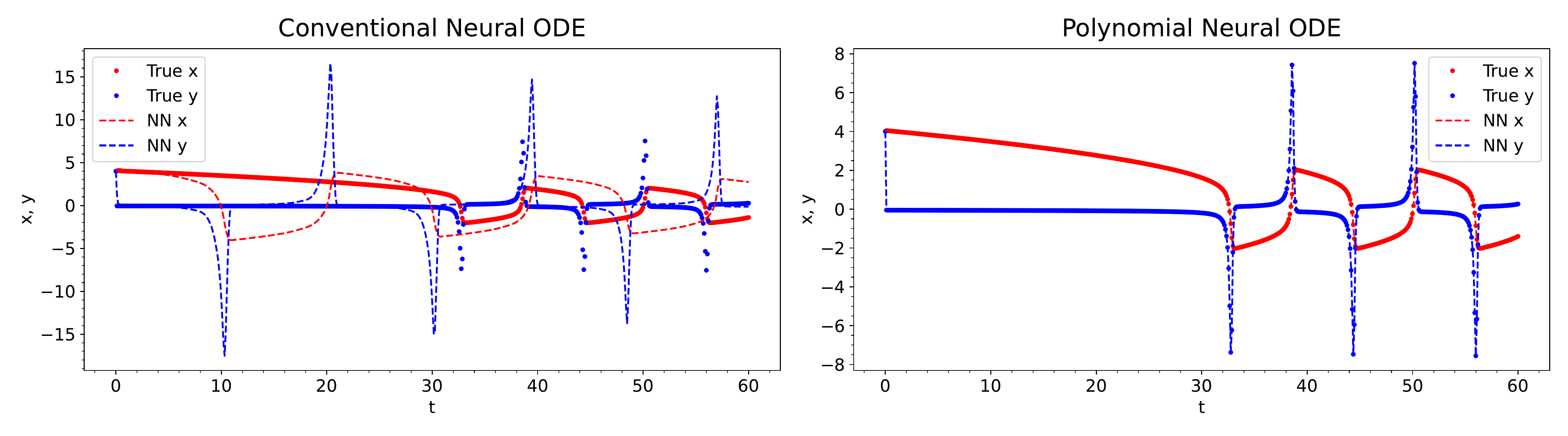}
    \caption{To test the ability of the neural ODEs to learn the limit cycle of the Van der Pol oscillator, the various models were integrated forward in time from initial conditions $x=4$ and $y=\frac{dx}{dt}=4$.  The conventional neural ODE with tanh activation functions did not learn the limit cycle from the data, whereas the third order polynomial neural ODE successfully did.}
    \label{fig:PiNet-Van-der-Pol-LimitCycle}
\end{figure}

\begin{figure}
    \centering
    \includegraphics[width=\textwidth]{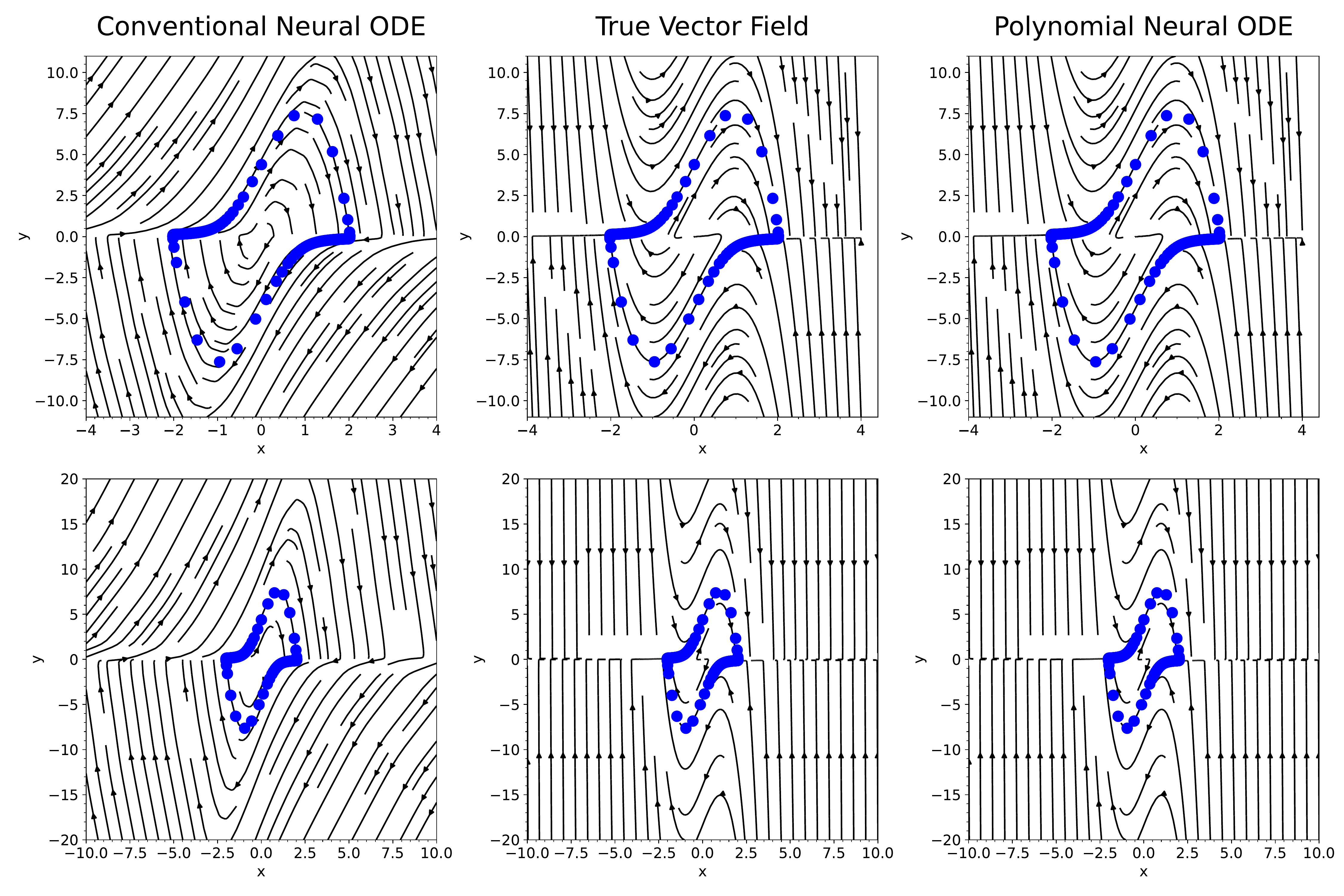}
    \caption{The vector fields of the time derivatives of variables x and y are plotted for the conventional neural ODE with tanh activations (left), the true ODE equation (middle), and the  third order $\pi$-net V1 polynomial neural ode (right). The top row shows the entire training region, while the bottom row shows the region outside of the training region.  The blue circles mark the training data.}
    \label{fig:PiNet-Van-der-Pol-VectorField}
\end{figure}

Following training, the performance of the neural networks were tested by: (1) integrating the neural ODEs from the training data's initial conditions up to a time value of 80 and (2) plotting the vector fields of the time derivatives of the variables x and y.  As shown in Fig \ref{fig:PiNet-Van-der-Pol-Full} and \ref{fig:PiNet-Van-der-Pol-VectorField}, the third order $\pi$-net polynomial neural ODE was able to successfully learn the dynamics of the system.  When the equations were recovered from the polynomial neural ODEs, as shown in Fig \ref{fig:PiNet-Van-der-Pol-Full}, the coefficients of the original Van der Pol ODE model were successfully recovered to 7 significant digits.

The conventional neural ODE with tanh activation functions was able to correctly predict the dynamical system's trajectory starting at the training data's initial conditions, as shown in Fig \ref{fig:PiNet-Van-der-Pol-Full}; however, the vector fields in Fig \ref{fig:PiNet-Van-der-Pol-VectorField} demonstrate that it learned slightly different dynamics from the original Van der Pol ODE the data was generated from.  The vector field of the traditional neural ODE does not show the limit cycle found in the Van der Pol ODE.  We can show this by integrating the neural ODEs and Van der Pol ODE forward in time starting at a different set of initial conditions.  Fig \ref{fig:PiNet-Van-der-Pol-LimitCycle} integrates the ODEs forward in time starting at $x=4$ and $y=\frac{dx}{dt}=4$.  It can be seen that the conventional neural ODE fails to learn the limit cycle, whereas the polynomial neural ODE does not have the same issue.

\FloatBarrier
\subsection{Learning Other Types of Equations with Polynomial Neural ODEs}

All of our previous demonstrations were on ODEs with equations defined by polynomials, so our last demonstration is on a dynamical system which is not a polynomial:

\begin{eqnarray} \label{eq:sincos}
    \frac{dx}{dt} &=& \cos(y)  \nonumber \\
    \frac{dy}{dt} &=&  - \sin(x). 
\end{eqnarray}

\noindent with initial conditions $x=0.5$ and $y=1$.  Since the problem is nonstiff, we generate our training data by integrating the IVP with SciPy and torchdiffeq using DOPRI5, a fourth order embedded method in the Runge–Kutta family of ODE solvers, with the default settings at 200 points uniformly spaced in time between 0 and 40 \citep{2020SciPy-NMeth, Chen_torchdiffeq_2021, dopriref}. As discussed in the methods section, we batch our data into 199 training samples consisting of IVPs between 2 adjacent known data points, and simultaneously solve the 199 IVPs during each epoch using the fourth order explicit Runge–Kutta–Fehlberg method \citep{fehlberg1968classical}.  The following neural networks were trained using the training data: (1) a conventional neural network with 5 layers consisting of 2x50x50x50x2 neurons in each layer with tanh activation functions, and (2) separate $\pi$-net V1 polynomial neural networks of degrees 4, 5, 6, and 15.  

\begin{figure}[h]
    \centering
    \includegraphics[width=\textwidth]{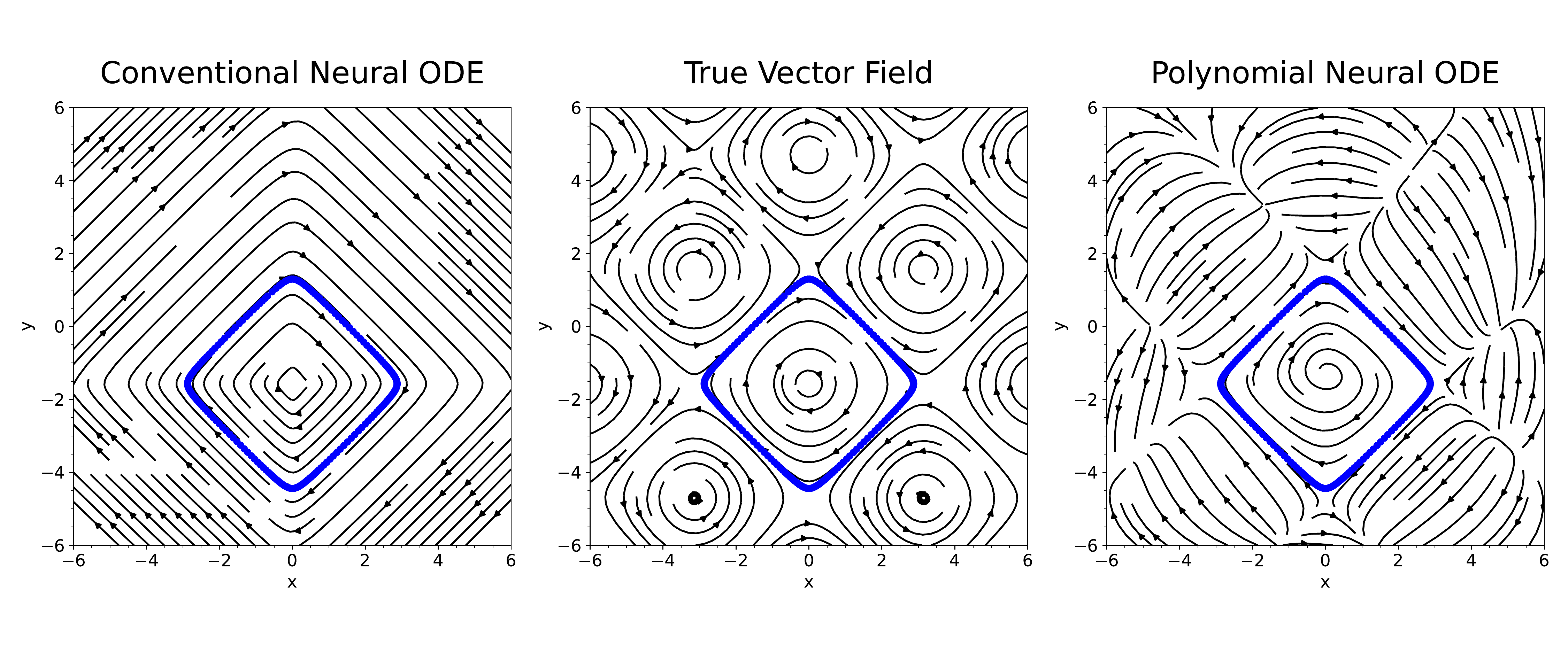}
    \caption{The vector fields of the time derivatives of variables x and y are plotted for the conventional neural ODE with hidden layers 2x50x50x50x2 and tanh activations (left), the true ODE equation (middle), and the sixth order $\pi$-net V1 polynomial neural ode (right).  The blue circles mark the training data.}
    \label{fig:PiNet-sincos-VectorField}
\end{figure}

Following training, the performance of the neural networks were tested by plotting the vector fields of the time derivatives of the variables x and y.  As shown in Fig \ref{fig:PiNet-sincos-VectorField}, the polynomial neural ODE was able to learn a better approximation to the true vector field than the conventional neural ODE.  While the polynomial neural ODE does not learn a perfect approximation to the field, it is better able to predict a rough approximation of the structure of the true vector field, including where additional spirals in the vector field would be, whereas the conventional neural ODE only learned one spiral.

\FloatBarrier
\section{Discussion/Conclusion}

This work introduced (1) symbolic regression with deep polynomial neural networks and (2) the polynomial neural ODE, which is a deep polynomial neural network implemented into the neural ODE framework.  Deep polynomial neural networks directly output a polynomial transformation of the input, which makes them directly interpretable.  We are the first to show that symbolic computing can be used to directly recover a symbolic representation of deep polynomial neural networks.  We demonstrated successful symbolic regression with a deep polynomial neural network on data generated from a univariate fourth order polynomial.  We also successfully demonstrated symbolic regression of dynamical systems governed by ODEs with the polynomial neural ODE on data from the Lotka-Volterra deterministic oscillator, damped oscillatory system, and Van der Pol deterministic oscillator.  

We understand that real world experimental data will be noisy, but we have chosen to devote an analysis on experimental noise to a future study.  We have a follow up paper on the data requirements for training conventional and polynomial neural ODEs such as noise, sampling frequency, and training size.  Additionally, most chemical kinetics systems are stiff ODEs \citep{doi:10.1137/1021001}, which arise from reaction rates that differ by many orders of magnitude, so neural ODEs need to be able to handle stiffness for this application space.  This work only shows examples for nonstiff ODEs.  Stiff ODEs require special treatment \citep{stiff_neural_ode}, which we will show in a follow up paper.

Traditional deep learning approaches from the field of computer science can be used on scientific problems, but this paper makes the case for developing deep learning techniques specifically tailored for scientific applications.  Rather than using "black box" data-driven approaches to describe physical phenomenon, we should be creating a suite of mechanistic data-driven approaches.  The polynomial neural ODE is one such approach; however, mathematical models usually have additional types of functions such as trigonometric functions and exponential functions.  The scientific machine learning community will need to work on developing more interpretable neural network architectures including more complicated functions such as these.

\section{Acknowledgements}

The authors acknowledge research funding from National Institutes of Health (NIH) NIBIB Award No. 2-R01-EB014877-04A1. Use was made of computational facilities purchased with funds from the National Science Foundation (CNS-1725797) and administered by the Center for Scientific Computing (CSC). The CSC is supported by the California NanoSystems Institute and the Materials Research Science and Engineering Center (MRSEC; NSF DMR 1720256) at UC Santa Barbara.  

This work was supported in part by NSF awards CNS-1730158, ACI-1540112, ACI-1541349, OAC-1826967, OAC-2112167, CNS-2120019, the University of California Office of the President, and the University of California San Diego's California Institute for Telecommunications and Information Technology/Qualcomm Institute. Thanks to CENIC for the 100Gbps networks.  The content of the information does not necessarily reflect the position or the policy of the funding agencies, and no official endorsement should be inferred.

\FloatBarrier
\bibliography{main.bib}

\end{document}